\documentclass[a4paper, 10pt, conference]{ieeeconf}

\IEEEoverridecommandlockouts
\usepackage{graphicx}
\usepackage{eso-pic}
\newif\ifarxiv
\arxivtrue 
\usepackage{amsmath,amssymb}
\usepackage{booktabs}
\usepackage{algorithm}
\usepackage{algorithmic}

\title{\LARGE \bf
Targeted Modality Dropout for Real-Robot Manipulation \\
Robust to Intermittent Vision Loss
}

\author{Genki Shikada$^{1,2}$, Kazuki Osamura$^{1}$, Masaru Ide$^{1}$, Tetsuya Ogata$^{2,3}$, and Kanata Suzuki$^{1}$%
\thanks{$^{1}$Fujitsu Limited, Kanagawa 211-8588, Japan. $^{2}$Faculty of Science and Engineering, Waseda University, Tokyo 169-8050, Japan. $^{3}$National Institute of Advanced Industrial Science and Technology, Tokyo 100-8921, Japan.
  {\tt\small shikada.genki@fujitsu.com}}%
}

\begin{document}

\ifarxiv
\AddToShipoutPictureFG*{\AtPageUpperLeft{%
  \put(\LenToUnit{0.5\paperwidth},\LenToUnit{-1.4cm}){%
    \makebox(0,0)[c]{\parbox{0.86\paperwidth}{\centering\footnotesize
\copyright{} 2027 IEEE.  Personal use of this material is permitted.  Permission from IEEE must be obtained for all other uses, in any current or future media, including reprinting/republishing this material for advertising or promotional purposes, creating new collective works, for resale or redistribution to servers or lists, or reuse of any copyrighted component of this work in other works.
    }}}}}
\fi
\maketitle
\thispagestyle{empty}
\pagestyle{empty}

\begin{abstract}
Imitation learning policies that integrate multiple sensory modalities are prone to overreliance on a dominant modality, such as vision, during training, which can disrupt policy execution when that modality is lost at inference time.
In this paper, we introduce Targeted Modality Dropout (TMD), in which the dependence on each modality is estimated using attention and the most dominant modality is selectively dropped during training.
This is combined with entropy regularization over the dependence distribution.
We evaluate six tasks on a real bimanual manipulator, simulating intermittent vision loss by jointly masking all vision tokens with 10\% probability per inference chunk.
Baseline successes decrease from 14/30 under normal conditions to 2/30 under vision loss, whereas TMD maintains an aggregate count of 15/30.
TMD also exhibits smaller changes in action commands when chunks generated under vision loss arrive.
In contrast, a conventional dropout that selects the dropped modality at random, without the entropy regularization, fails on many tasks even without vision loss.
\end{abstract}

\section{INTRODUCTION}

In imitation learning, a manipulation policy is developed directly from demonstrations, without a tailored reward function or controller. A human operator teleoperates the robot to perform a task while the camera images, the robot's own state and the commanded actions are recorded. The policy is then trained via supervised learning to predict the actions from the observations. At run time the task is executed in a closed loop; thus, each step is driven by the sensor inputs available at that moment. Recent studies have increasingly adopted vision-language-action models, which encode images and language instructions using pretrained foundation models and generate actions from the resulting representation~\cite{tvfdit2026, zhang2025tavla}. In contact-rich manipulation, vision alone does not reveal how much force is applied or whether contact has been established; thus, integrating several modalities such as vision, language, and force is intuitive~\cite{famil2026}.

\begin{figure}[t]
\centering
\includegraphics[width=\columnwidth]{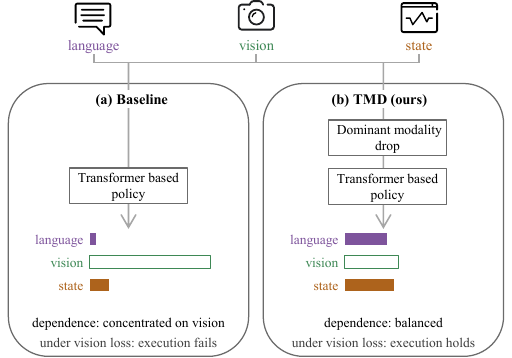}
\caption{Overview of the problem and of the proposed method. Three modalities feed one policy. (a) Under ordinary training the dependence, estimated from attention, concentrates on vision. (b) TMD drops whichever modality is dominant during training, and the dependence balances. The bars are the estimated dependence at the end of training, with the vision bar left unfilled to mark what inference-time loss removes.}
\label{fig:teaser}
\end{figure}

However, when a policy is trained by integrating multiple modalities, potential overreliance on a particular modality may lead to underuse of the others~\cite{wu2022greedy}. Vision, in particular, tends to become dominant because it is information-rich; consequently, vision loss at inference time can severely impair policy execution. Camera failures and occlusions do occur in real-world deployment, making overreliance on a particular modality a practical robustness concern, as illustrated in Fig.~\ref{fig:teaser}.

Modality dropout can intervene in this bias by stochastically removing modalities during training~\cite{neverova2016moddrop, liu2017sensordropout}. Randomly removing modalities, however, does not account for which modality is dominant and thus does not correct the bias itself. In this study we propose Targeted Modality Dropout (TMD), which estimates each modality's dependence from attention and drops the modality with the largest dependence as a target. We further introduce entropy regularization over the dependence distribution to alleviate concentration on any single modality.

To date, robustness to missing modalities has been assessed mainly using multimodal classification benchmarks or simulated manipulation. Situations where vision is lost while a real robot executes a task, i.e., robustness in a closed loop, have rarely been examined. In this study we apply TMD to a real bimanual manipulator and evaluate whether task execution is sustained when vision is stochastically lost at inference time. We also examine the effect of dependence-based targeted selection by comparing TMD with a conventional dropout in which the dropped modality is selected at random, referred to as Random Modality Dropout (RMD).

The contributions of this study are as follows:
\begin{enumerate}
\item We propose Targeted Modality Dropout (TMD), which drops the dominant modality as a target based on the dependence estimated from attention, together with entropy regularization.
\item Through real-robot closed-loop evaluation using a bimanual manipulator, we demonstrate that TMD sustains task execution under vision loss and that its action commands are disturbed less.
\item We show that RMD, which replaces the targeted selection with a random one and omits the entropy regularization, fails on many tasks even without vision loss.
\end{enumerate}

\section{RELATED WORK}

\subsection{Multimodal Imitation Learning for Manipulation}
In manipulation, the modalities received by a policy carry different types of information. Images provide the position and orientation of the objects, and a language instruction specifies the task to be performed. Force or tactile signals report the contact that images cannot capture~\cite{famil2026}. Transformer-based policies treat these inputs as a single token sequence to generate an action chunk~\cite{zhao2023aloha}. In vision-language-action models, pretrained encoders provide the visual and language tokens~\cite{tvfdit2026, zhang2025tavla}. These inputs substitute for one another only partially. Thus, losing one of them at inference time does not uniformly degrade the performance of the policy; only what that input alone was supplying is lost. Nevertheless, the extent to which a policy relies on each input is determined implicitly through training, and such policies are typically developed under the assumption that all inputs are available.

\subsection{Controlling Modality Reliance}
When a policy is trained on multiple modalities, it may over-rely on a dominant one and fail to use the others~\cite{wu2022greedy}. Several lines of research have addressed this issue. Fusion-mechanism design suppresses the dominance of a particular modality through architectural choices~\cite{famil2026}; however, this approach is closely tied to the specific policy architecture. Phase-dependent weighting dynamically adjusts the contribution of each modality according to the task phase~\cite{ichiwara2023modality, msbot2024}, but presupposes that all modalities remain available. Visual degradation during training reduces the policy's reliance on vision~\cite{liu2025factr} but does not examine which modality is dominant. Gradient rebalancing modulates the gradient of each modality based on its contribution at each iteration~\cite{peng2022ogm, li2023agm} but does not train the policy to handle the absence of a modality. Modality dropout stochastically drops modalities during training~\cite{neverova2016moddrop, liu2017sensordropout}, but randomly selecting the dropped modality does not account for modality dominance. In a setting closer to ours, modality dropout has been used to keep a network from ignoring the weaker of two modalities in audiovisual facial animation~\cite{abdelaziz2020modalitydropout}, with the probability of dropping each modality tuned in advance based on prior knowledge of the dominant modality rather than on a quantity estimated during training. Because this study targets robustness to the loss of the dominant modality, we identify the dominant modality using attention and selectively drop it during training (Section~\ref{sec:method}).

\subsection{Evaluation under Missing Modalities}
Most of the methods described above are evaluated under normal operating conditions that do not involve complete modality loss. Research on visuotactile fusion addresses vision-constrained conditions, but remains limited to situations where vision is only partially restricted~\cite{gelfusion2025}. Robustness to missing modalities has been studied mainly outside real-robot closed-loop settings: on multimodal classification benchmarks~\cite{ma2022missing} or simulated manipulation~\cite{retrieval2026}. The scenario in which vision is stochastically lost while a real robot executes a task in closed loop has rarely been examined. Unlike existing studies, this study focuses on evaluating real-robot task execution under inference-time vision loss.

\section{PROPOSED METHOD}
\label{sec:method}

\begin{figure*}[t]
\centering
\includegraphics[width=\textwidth]{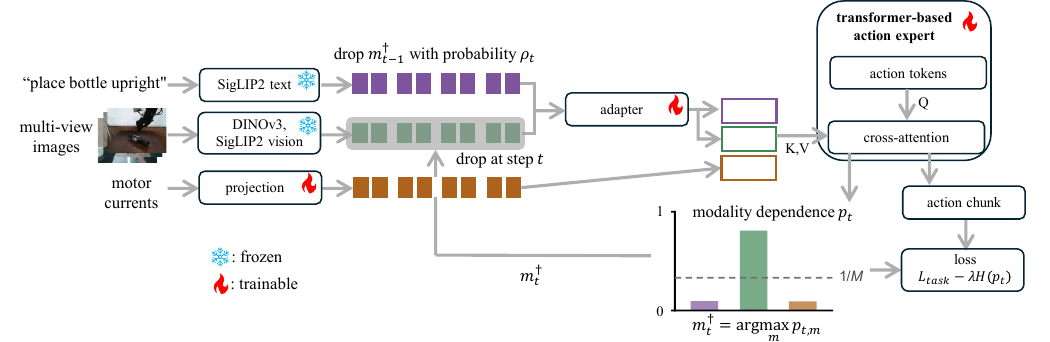}
\caption{Overview of the proposed method during training. Language and vision tokens are encoded by frozen pretrained encoders and integrated by a trainable adapter, while state tokens are projected and joined directly, so that the key/value memory of the action expert contains the tokens of all three modalities. Because the adapter attends across the language and vision tokens, each of its output tokens carries information from both, so the dependence on these two modalities is not measured on fully independent representations. The dependence $p_t$ drives two mechanisms: the tokens of $m^{\dagger}_{t-1}$ are zeroed in a subset of samples, and the entropy of $p_t$ enters the loss. In the step drawn here, vision is the target, so its token row is shaded and its tokens are dimmed. See Algorithm~\ref{alg:tmd} for the order of operations.}
\label{fig:model}
\end{figure*}
\subsection{Problem Formulation and Notation}
The policy considered in this study is a Transformer-based vision-language-action model that integrates input tokens from multiple modalities via cross-attention and generates an action chunk~\cite{zhao2023aloha}. The input modalities are language (language instructions), vision (images), and state (motor currents), with $\mathcal{M} = \{\mathrm{language}, \mathrm{vision}, \mathrm{state}\}$ and $M = |\mathcal{M}| = 3$. Here, state refers to motor currents, which serve as an effective signal for reflecting contact states without using an external force sensor~\cite{zhang2025tavla}.

Fig.~\ref{fig:model} provides an overview of the proposed method. Based on the dependence on each modality estimated from cross-attention, the proposed method suppresses the dominant modality as a target. Hereafter, we describe the estimation of dependence (Section III-B), and define the targeted modality dropout (Section III-C) and entropy regularization (Section III-D).

\subsection{Estimation of Modality Dependence}
We first estimate the dependence $p_{t,m}$ on each modality from the cross-attention weights of the Transformer. Let $\ell = 1, \ldots, L$ be the layer index, $h = 1, \ldots, H$ the head index, $\mathcal{Q}$ the set of queries, and $\mathcal{T}_m$ the set of key/value tokens belonging to modality $m$. Let $\alpha^{(\ell,h)}_{t,qk}$ denote the attention weight from query $q$ to token $k$. The layer- and head-averaged weight $\bar{\alpha}_{t,qk}$ and the dependence $p_{t,m}$ are given by
\begin{equation}
\begin{aligned}
\bar{\alpha}_{t,qk} &= \frac{1}{LH}\sum_{\ell=1}^{L}\sum_{h=1}^{H}\alpha^{(\ell,h)}_{t,qk}, \\
p_{t,m} &= \frac{\sum_{q\in\mathcal{Q}}\sum_{k\in\mathcal{T}_m}\bar{\alpha}_{t,qk}}{\sum_{m',q,k}\bar{\alpha}_{t,qk}}.
\end{aligned}
\end{equation}
The dependence $p_{t,m}$ represents the relative magnitude of dependence on modality $m$, and satisfies $\sum_{m\in\mathcal{M}} p_{t,m} = 1$.

\subsection{Targeted Modality Dropout (TMD)}
\label{sec:tmd}

\begin{algorithm}[t]
\caption{One training step with Targeted Modality Dropout}
\label{alg:tmd}
\begin{algorithmic}[1]
\REQUIRE minibatch $\mathcal{B}$, step $t$, target $m^{\dagger}_{t-1}$ from the previous step
\STATE $\rho_t \leftarrow \mathrm{schedule}(t)$ \hfill $\triangleright$ $0 \rightarrow \rho_{\max} \rightarrow 0$
\FORALL{samples $b \in \mathcal{B}$}
  \STATE draw $\xi_{t,b} \sim \mathrm{Uniform}(0,1)$
  \STATE zero all tokens of $m^{\dagger}_{t-1}$ in sample $b$ if $\xi_{t,b} < \rho_t$ \hfill $\triangleright$ Eq.~(2)
\ENDFOR
\STATE run the encoder and the action expert on the masked minibatch
\STATE $p_t \leftarrow$ dependence aggregated from the cross-attention of the action expert \hfill $\triangleright$ Eq.~(1)
\STATE $\mathcal{L} \leftarrow \mathcal{L}_{\mathrm{task}} - \lambda H(p_t)$ \hfill $\triangleright$ Eq.~(3), (4)
\STATE update the policy with $\nabla \mathcal{L}$
\STATE $m^{\dagger}_t \leftarrow \arg\max_{m\in\mathcal{M}} p_{t,m}$ \hfill $\triangleright$ used at the next step
\end{algorithmic}
\end{algorithm}

Based on the estimated dependence, we select the dominant modality during training and stochastically drop that modality. Let $m^{\dagger}_t = \arg\max_{m\in\mathcal{M}} p_{t,m}$ be the modality with the largest dependence.

Dropout is applied independently to each sample. For each sample $b$ in a batch, we independently draw a uniform random variable $\xi_{t,b} \sim \mathrm{Uniform}(0,1)$. When $\xi_{t,b} < \rho_t$, we zero out all tokens of the modality $m^{\dagger}_{t-1}$ selected at the preceding step in sample $b$. Let $\mathcal{T}_m$ be the set of tokens of modality $m$, and let $x^{(m)}_{t,b,k}$ be the token at position $k$ of modality $m$ in sample $b$, as produced by the encoder of that modality. The masked token is $\tilde{x}^{(m)}_{t,b,k} = r^{(m)}_{t,b}\, x^{(m)}_{t,b,k}$, wherein the mask coefficient $r^{(m)}_{t,b}$ is defined by
\begin{equation}
\begin{aligned}
r^{(m)}_{t,b} &= 1 - \mathbf{1}[m = m^{\dagger}_{t-1}]\,\mathbf{1}[\xi_{t,b} < \rho_t], \\
\xi_{t,b} &\sim \mathrm{Uniform}(0,1).
\end{aligned}
\end{equation}
The mask coefficient $r^{(m)}_{t,b}$ does not depend on the token index $k$ but only on the sample $b$. Therefore, in a selected sample, all tokens of $m^{\dagger}_{t-1}$ are zeroed out simultaneously, blocking information at the modality level.

The dropout probability $\rho_t$ is not constant throughout training but follows a schedule. In the early stage of training, $\rho_t = 0$. This probability is then ramped up to an upper bound $\rho_{\max}$ partway through training. In the late stage of training, $\rho_t$ is decayed back to $0$, so that training ends in a state without dropout. Algorithm~\ref{alg:tmd} summarizes one training step. As indicated by the subscripts in (2), the estimation of the target and the application of the mask are separated by one step: the mask at step $t$ uses $m^{\dagger}_{t-1}$, while the attention at step $t$ yields $m^{\dagger}_t$, which is used at step $t+1$.

\subsection{Entropy Regularization}
To suppress the dependence bias, we introduce entropy regularization. We use the form of the confidence penalty~\cite{pereyra2017regularizing}, which adds the negative entropy of a distribution to the training loss, preventing that distribution from becoming peaked. In that work the penalty is applied to the output distribution over classes; here we apply it to the dependence distribution over modalities. The entropy of the dependence distribution $p_t = (p_{t,m})_{m\in\mathcal{M}}$ is defined by
\begin{equation}
H(p_t) = -\sum_{m\in\mathcal{M}} p_{t,m} \log(p_{t,m} + \varepsilon),
\end{equation}
where $\varepsilon > 0$ is a small constant for numerical stability, set to $\varepsilon = 1\times 10^{-8}$ in this study. Maximizing $H(p_t)$ corresponds to bringing the dependence distribution closer to a uniform distribution, suppressing the concentration of dependence on a particular modality and promoting balance toward $1/M \simeq 0.33$. The total training loss is defined, relative to the task loss $\mathcal{L}_{\mathrm{task}}$, as
\begin{equation}
\mathcal{L} = \mathcal{L}_{\mathrm{task}} - \lambda H(p_t),
\end{equation}
where $\lambda \geq 0$ is a weight that determines the strength of the regularization, set to $\lambda = 0.001$ in this study.

\begin{figure*}[t]
\centering
\includegraphics[width=\textwidth]{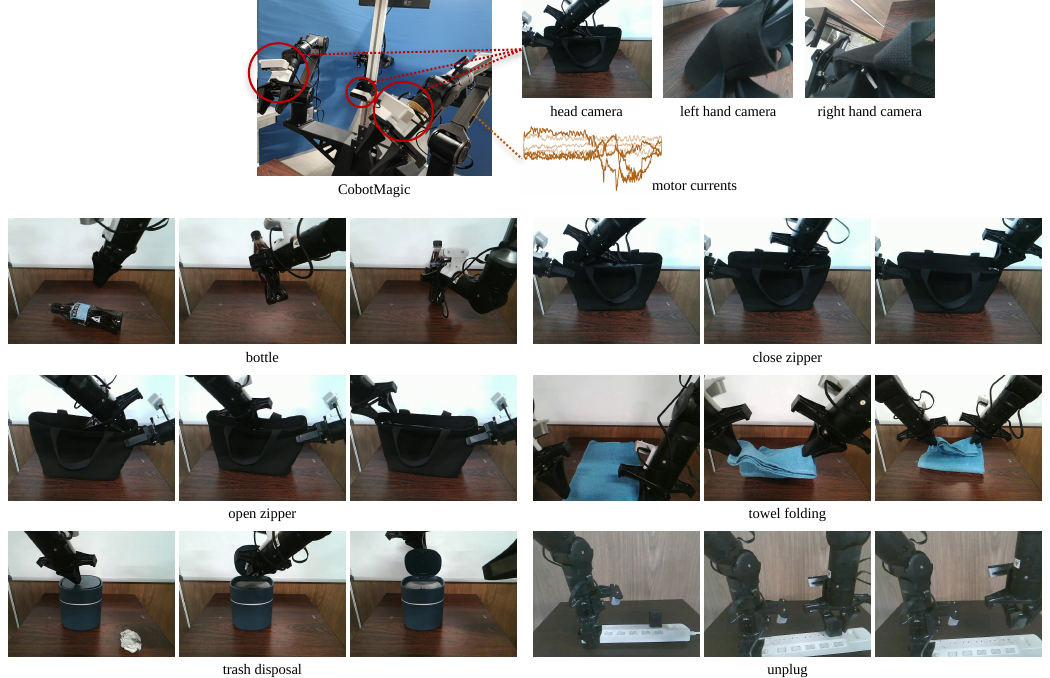}
\caption{Top: modalities the policy receives from the bimanual manipulator CobotMagic: images from the head and the left- and right-hand cameras, taken at the moment when the slider of the fastener is grasped to close the zipper, and the motor currents. Bottom: six tasks used for real-world evaluation: bottle, close zipper, open zipper, towel folding, trash disposal, and unplug.}
\label{fig:tasks}
\end{figure*}

\newpage
\section{EXPERIMENTS}
This section presents the experimental conditions, evaluation conditions, results, and discussion.

\subsection{Experimental Setup}

\subsubsection{Implementation}
We use a bimanual manipulator, CobotMagic, as the real robot (Fig.~\ref{fig:tasks}). The action is 14-dimensional in total for both arms, concatenating the 6-dimensional joint angles and 1-dimensional gripper of each arm. The inputs are RGB cameras from three viewpoints---head, left-hand, and right-hand (vision); the motor currents of the left and right arms (state); and language instructions (language) (Section III-A). The policy is a vision-language-action model based on TVF-DiT~\cite{tvfdit2026}. Using a trainable adapter, the visual encoder integrates the features of a pretrained visual foundation model, DINOv3 (ViT-S/16)~\cite{simeoni2025dinov3}, and a vision-language model, SigLIP2~\cite{tschannen2025siglip2}. Following the use of diffusion models as action decoders for visuomotor manipulation~\cite{chi2025diffusion}, the action decoder is implemented as a Diffusion Transformer~\cite{peebles2023dit} trained with Conditional Flow Matching (CFM)~\cite{lipman2023flow}. The total number of parameters in the policy is approximately 471 million. DINOv3 and SigLIP2 are used with their pretrained weights frozen. The dependence $p_{t,m}$ in Eq.~(1) is aggregated from the cross-attention weights of the action expert, whose key/value memory contains the tokens of all three modalities. For training, we use 100 demonstrations per task, a batch size of 32 (with gradient accumulation of 2), and the AdamW optimizer~\cite{loshchilov2019adamw} (learning rate $1\times10^{-4}$). Training is performed for 300k steps, and that checkpoint is used for all evaluations. Following the schedule in Section III-C, the dropout probability $\rho$ of the targeted modality dropout reaches $\rho_{\max}=0.6$ at 60k steps and decays to $0$ from 240k to 300k steps.

\subsubsection{Comparison Conditions}
We compare the following three conditions:
\begin{itemize}
\item \textbf{Baseline}: ordinary training without dropout or entropy regularization.
\item \textbf{TMD}: the proposed method, combining the targeted dropout of Section III-C with the entropy regularization of Section III-D.
\item \textbf{RMD}: a control that randomizes the selection of the modality dropped in TMD. At each step, the dropped modality is chosen by uniform random sampling from the candidates rather than as the dominant modality, and no entropy regularization is applied. That is, RMD and TMD share the same dropout mechanism, but the way the dropped modality is selected differs. In terms of Algorithm~\ref{alg:tmd}, RMD replaces line~10 with a uniform random choice and omits the $-\lambda H(p_t)$ term in line~8.
\end{itemize}

\subsubsection{Evaluation}
We use the following six tasks for real-robot evaluation:
\begin{itemize}
\item \textbf{bottle}: grasp a fallen bottle with the right gripper and stand it upright.
\item \textbf{close zipper}: hold the bag and close its initially open zipper.
\item \textbf{open zipper}: hold the bag and open its initially closed zipper.
\item \textbf{towel folding}: fold a spread-out towel in half.
\item \textbf{trash disposal}: push the bin lid open, pick up the paper trash, place it in the bin, and close the lid.
\item \textbf{unplug}: hold the socket in place and pull out the plug.
\end{itemize}

Each trial is limited to 3,000 action-command steps. A trial is successful only if the entire task is completed within this limit; partial completion, such as an incompletely closed zipper or a partially folded towel, is counted as a failure. For each task and each condition, we measure the number of successes / number of trials ($N=5$ per condition). Evaluation is performed under two settings: a normal setting in which all modalities are available throughout each trial, and a vision-loss setting that simulates intermittent unavailability of all camera inputs. In the vision-loss setting, all vision tokens from the three cameras are masked simultaneously with probability $10\%$ at each inference chunk, independently of other chunks, using the same token-masking operation as in Section~\ref{sec:tmd}.

We consider temporary interruptions of image delivery to the policy as a motivating example. This masking provides a controlled test of complete visual-input unavailability at chunk generation. Partial object occlusions and the temporal structure of specific sensor failures are not modeled.

\begin{figure}[t]
\centering
\includegraphics[width=\columnwidth]{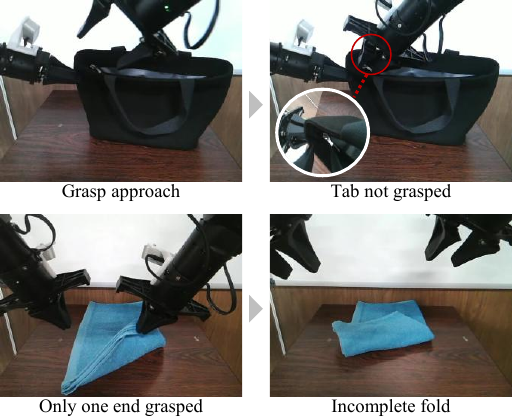}
\caption{Representative TMD failures under the normal setting. Top: close zipper. The robot approaches the zipper pull tab but fails to grasp it, leaving the zipper open. The inset provides an additional close-up view of the ungrasped tab. Bottom: towel folding. Only one end of the towel is grasped, and the fold remains incomplete. Images in each row are ordered chronologically from left to right.}
\label{fig:failures}
\end{figure}

\begin{figure*}[t]
\centering
\includegraphics[width=\textwidth]{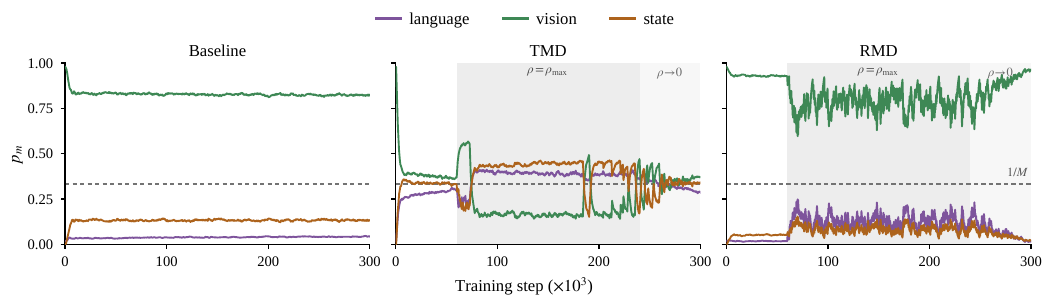}
\caption{Trajectory of the modality dependence $p_{t,m}$ during training, for each condition. The three modalities are overlaid in each panel; curves are exponential moving averages. The dashed horizontal line marks the uniform level $1/M \simeq 0.33$, and the shaded band marks the interval in which the dropout probability $\rho_t$ is nonzero (Baseline applies no dropout).}
\label{fig:dependency}
\end{figure*}

\begin{figure}[t]
\centering
\includegraphics[width=\columnwidth]{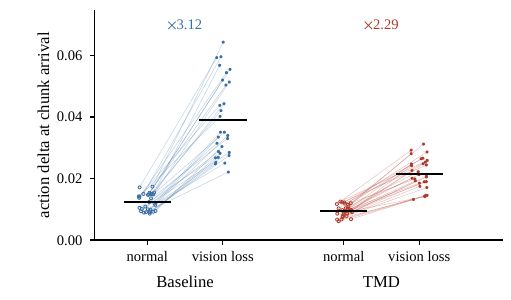}
\caption{Action delta at chunk arrival within the $30$ vision-loss trials per method. The labels ``normal'' and ``vision loss'' denote chunks generated without and with vision-token masking, respectively, within the same trial. Open and filled markers indicate the corresponding within-trial means and are connected by a line; horizontal bars are means across trials. The mean within-trial ratio (masked/unmasked), marked by $\times$, is $3.12$ for Baseline versus $2.29$ for TMD. Action delta is the sum over the two arms of the L2 norm of the change in the commanded action vector between adjacent steps; the action vector comprises the six joint angles and the gripper command and is reported in arbitrary units.}
\label{fig:actiondelta}
\end{figure}

\subsection{Real-Robot Evaluation}
This section evaluates the robustness to vision loss on the real robot in terms of task success rate.

\subsubsection{Normal Setting}
Table~\ref{tab:normal} shows the number of successes / number of trials on the real robot under the normal setting in which all modalities are available ($N=5$ per condition). Baseline and TMD have comparable total success counts, 14/30 and 15/30 respectively, with their relative advantage varying by task.

For close zipper, TMD failed to grasp the zipper pull tab in all five trials. Baseline completed the task in three trials: one with a successful first grasp attempt and two after repeated grasp attempts within the same trial. In the two unsuccessful Baseline trials, the robot failed to grasp the tab in one and released it before completing zipper closure in the other.

For towel folding, TMD completed three of the five trials without retries. In its two unsuccessful trials, only one end of the towel was grasped. Baseline completed four of the five trials without retries; in the remaining trial, the grasp on one end of the towel was incomplete.

Fig.~\ref{fig:failures} shows representative TMD failures in these two tasks under the normal setting.

In contrast, RMD achieves only 4/30 in total, with three of its four successes on the unplug task. Because RMD does not succeed on many tasks even under the normal setting with all modalities provided, we restrict the subsequent vision-loss evaluation to Baseline and TMD.

\begin{table}[t]
\caption{Number of successes / number of trials on the real robot under the normal setting ($N=5$ per condition).}
\label{tab:normal}
\centering
\begin{tabular}{lccc}
\toprule
Task & Baseline & TMD & RMD \\
\midrule
bottle & 2/5 & 5/5 & 0/5 \\
close zipper & 3/5 & 0/5 & 0/5 \\
open zipper & 1/5 & 2/5 & 0/5 \\
towel folding & 4/5 & 3/5 & 1/5 \\
trash disposal & 1/5 & 2/5 & 0/5 \\
unplug & 3/5 & 3/5 & 3/5 \\
\midrule
Total & 14/30 & 15/30 & 4/30 \\
\bottomrule
\end{tabular}
\end{table}

\subsubsection{Vision-Loss Setting}
Table~\ref{tab:vloss} shows the number of successes / number of trials under the setting where all camera images are lost with probability 10\% at each inference chunk ($N=5$ per condition). $\Delta$ is the change in the success count relative to the normal setting (Table~\ref{tab:normal}). For Baseline, the success count drops from 14/30 to 2/30 in total (total $\Delta=-12$), decreasing on every task, indicating vulnerability to vision loss. In contrast, the success count of TMD remains unchanged at 15/30 in total (total $\Delta=0$). Moreover, the success count drop for TMD is smaller than that for Baseline. However, this total $\Delta=0$ results from mutual cancellation of the task-wise increases and decreases: while towel folding and trash disposal decrease, unplug increases.

During the vision-loss trials, we qualitatively observed abrupt, jump-like end-effector motions. Such disturbances may have disrupted task execution and contributed to the task-wise reductions in success counts. This interpretation is consistent with the larger action deltas for masked chunks than for unmasked chunks in Fig.~\ref{fig:actiondelta}. The action delta quantifies changes in action commands, and this correspondence does not establish the cause of each individual trial failure.

\begin{table}[t]
\caption{Number of successes / number of trials on the real robot under vision loss ($N=5$ per condition). $\Delta$ is the change from the normal setting.}
\label{tab:vloss}
\centering
\begin{tabular}{lcc}
\toprule
Task & Baseline ($\Delta$) & TMD ($\Delta$) \\
\midrule
bottle & 1/5 ($-1$) & 5/5 ($0$) \\
close zipper & 0/5 ($-3$) & 0/5 ($0$) \\
open zipper & 0/5 ($-1$) & 2/5 ($0$) \\
towel folding & 1/5 ($-3$) & 2/5 ($-1$) \\
trash disposal & 0/5 ($-1$) & 1/5 ($-1$) \\
unplug & 0/5 ($-3$) & 5/5 ($+2$) \\
\midrule
Total & 2/30 ($-12$) & 15/30 ($0$) \\
\bottomrule
\end{tabular}
\end{table}

\subsection{Analysis}
This section analyzes the mechanism and the evidence supporting the robustness observed in the real-robot evaluation. We first show that modality dependence is corrected during training and then examine the continuity of actions under vision loss.

\subsubsection{Correction of Modality Dependence during Training}
Fig.~\ref{fig:dependency} shows the trajectory of the modality dependence $p_{t,m}$ during training for each condition. For Baseline, because no control is applied, the dependence on vision becomes dominant while the dependence on language and state remains low. In contrast, for TMD, the dependence on vision decreases and the dependences of the three modalities become more balanced, thereby mitigating the bias toward a single modality. For RMD, however, because the dropped modality is selected at random, the dependence does not become more balanced and the dependence on vision remains dominant throughout. In other words, TMD, which selects the target based on dependence, mitigates the dependence bias, whereas RMD, which selects the target at random, does not correct it.

Three observations add detail to this picture. First, the two components of TMD act in different intervals. Before the dropout probability is ramped up, entropy regularization is the only mechanism in effect, and it already brings the dependence on language, vision and state to $0.29$, $0.37$ and $0.34$ respectively. When dropout is applied, the dependence on vision declines to $0.19$, below the uniform level, but returns to $0.37$ once $\rho$ decays back to zero. Second, the target is not fixed to a single modality: over the interval in which dropout is applied, TMD selects vision as the target in $0.51$ of the steps, state in $0.45$ and language in $0.05$, whereas RMD selects each modality in $0.33$ of the steps by construction, suggesting that vision and state alternate as the target in TMD. Third, at the end of training the dependence in TMD remains close to balanced, at $0.29$, $0.37$ and $0.34$ for language, vision and state respectively, whereas in RMD the dependence on vision has returned to $0.95$. Because all evaluations use the checkpoint at the end of training, this last state is the one in which the policies are evaluated.

\subsubsection{Continuity of Actions under Vision Loss}
The success rate captures whether execution succeeds or fails, but it does not capture the effect of vision loss on the policy's action generation itself. We therefore analyze action delta at chunk arrival within the vision-loss trials. For each step, we compute the action delta as the sum over the two arms of the L2 norm of the change in the commanded action vector between adjacent steps. For each trial, we average this value separately over the steps at which chunks generated with vision-token masking arrive and the steps at which chunks generated without vision-token masking arrive.

Fig.~\ref{fig:actiondelta} shows these two within-trial means for each of the $30$ vision-loss trials per method. For both methods, the mean action delta at arrival is larger for masked chunks than for unmasked chunks. The mean within-trial ratio (masked/unmasked) is $3.12$ for Baseline and $2.29$ for TMD (Mann-Whitney $U$ test, $p = 1.6 \times 10^{-8}$, $n = 30$ per condition). In absolute terms, the mean action delta for masked chunks is $0.039$ for Baseline and $0.022$ for TMD, whereas the corresponding means for unmasked chunks are $0.012$ and $0.010$, respectively. The mean increase from the unmasked-chunk level is therefore $0.027$ for Baseline and $0.012$ for TMD. Because the unmasked-chunk means also differ between the methods, we report both the absolute action deltas and the within-trial ratios, with the latter expressing the increase relative to each trial's unmasked-chunk level.

\subsection{Discussion}
In the normal setting (Table~\ref{tab:normal}), TMD does not outperform Baseline on all tasks. The observed failure modes point to difficulties in acquiring or maintaining the required grasps, particularly in TMD's grasping of the zipper pull tab. Suppressing visual dependence may contribute to such difficulties in precise grasping, but the present observations do not establish this explanation. However, as shown in Table~\ref{tab:vloss}, although the success count of Baseline drops on many tasks under vision loss, the drop is smaller for TMD. Thus, the value of TMD lies not in improving the success rate under normal conditions but in robustness under vision loss.

The reason TMD can sustain execution under vision loss is considered to be that it corrected the dependence on vision toward balance during training (Section IV-C). Since the dependence on vision is suppressed, less of what the policy relies on is removed when vision is lost, which may make it easier to continue execution. Moreover, the fact that the action commands of TMD are disturbed less at the arrival of a chunk generated under vision loss (Section IV-C) shows that this robustness also manifests as stability of actions. However, this work does not directly demonstrate a causal relationship between the correction of dependence and the robustness, and these correspondences remain interpretations.

RMD, which selects the dropped modality at random, does not succeed on many tasks even under the normal setting (Table~\ref{tab:normal}). This suggests the importance of selecting the target based on dependence rather than selecting the dropped modality at random. However, because TMD and RMD in this study differ not only in the way the dropped modality is selected but also in the presence or absence of entropy regularization, the difference between them cannot be attributed solely to the selection method.

The real-robot evaluation is limited to $N=5$ trials per condition; thus, small differences in the success count may be subject to error. In fact, some changes are hard to attribute to vision loss making execution easier---for example, under vision loss, TMD's success count on unplug increases relative to the normal setting (Table~\ref{tab:vloss})---and the values for individual tasks are considered to include variation due to the number of trials. TMD is already trained with missing modalities, including vision, and the present experiments evaluate robustness to intermittent vision loss at inference. Robustness to sustained vision loss, including comparison with a policy trained specifically for persistent visual unavailability, is not evaluated in this study.

\section{CONCLUSION}
To address the bias toward a dominant modality in multimodal imitation learning, we introduced TMD, which selectively drops the dominant modality based on the dependence estimated using attention, together with entropy regularization. Through real-robot evaluation using a bimanual manipulator, CobotMagic, we showed that under vision loss the success rate of Baseline drops substantially, whereas TMD sustains execution. We also confirmed that, under vision loss, the action commands of TMD are less perturbed by the arrival of an affected chunk. In contrast, RMD, which selects the dropped modality at random and omits entropy regularization, fails on many tasks even without vision loss.

However, the number of trials used in the real-robot evaluation is limited, TMD and RMD differ in the selection rule and in the use of the regularization, and the evaluated vision loss is short-term. In future research, we will expand the trials and tasks, separate the selection rule from the entropy regularization, and evaluate sustained modality loss.

\section*{ACKNOWLEDGMENT}
This work was supported by JST PRESTO, Japan, Grant Number JPMJPR24T4.

\bibliographystyle{IEEEtran}
\bibliography{references}

\end{document}